\documentclass[conference]{IEEEtran}
\IEEEoverridecommandlockouts
\usepackage{cite}
\usepackage{amsmath,amssymb,amsfonts}
\usepackage{algorithmic}
\usepackage{graphicx}
\usepackage{textcomp}
\usepackage{multicol}
\usepackage{xcolor}
\usepackage{colortbl}
\usepackage{booktabs}
\usepackage{multirow}
\usepackage{url}
\usepackage{hyperref}
\usepackage{graphicx} %resizing table
\usepackage{stmaryrd} % Required for \llbracket and \rrbracket
\def\BibTeX{{\rm B\kern-.05em{\sc i\kern-.025em b}\kern-.08em
    T\kern-.1667em\lower.7ex\hbox{E}\kern-.125emX}}

\definecolor{Gray}{gray}{0.9}
\definecolor{syntheticcolor}{HTML}{E0FFFF} % Light Cyan for Synthetic
\definecolor{brightfieldcolor}{HTML}{F0FFF0} % Honeydew for Brightfield
\definecolor{fluorescencecolor}{HTML}{FFE4E1} % MistyRose for Fluorescence
\DeclareMathOperator*{\argmin}{argmin}
\begin{document}

\title{CellFMCount: A Fluorescence Microscopy Dataset, Benchmark, and Methods for Cell Counting\\
%{\footnotesize \textsuperscript{*}Note: Sub-titles are not captured for https://ieeexplore.ieee.org  and
%should not be used}
\thanks{This work is partially supported by the National Science Foundation under Grant No. 2152117. Any opinions, findings, and conclusions or recommendations expressed in this material are those of the author(s) and do not necessarily reflect the views of the National Science Foundation.}
}

\author{
\textbf{Abdurahman Ali Mohammed}$^{1,6}$, \textbf{Catherine Fonder}$^{2,3,5,6}$, \textbf{Ying Wei}$^{1,6}$, \\ \textbf{Wallapak Tavanapong}$^{1,6}$,
\textbf{Donald S. Sakaguchi}$^{2,3,4,5,6}$, \textbf{Qi Li}$^{1,6}$, \textbf{Surya K. Mallapragada}$^{2,3,4,5,6}$ \\
\textit{$^{1}$Department of Computer Science} \\
\textit{$^{2}$Department of Genetics, Development, and Cell Biology} \\
\textit{$^{3}$Molecular, Cellular, and Developmental Biology Program} \\
\textit{$^{4}$Neuroscience Program} \textit{$^{5}$Nanovaccine Institute} \\
\textit{$^{6}$Iowa State University, Ames, IA 50011} 
}
\maketitle

\begin{abstract}
Accurate cell counting is essential in various biomedical research and clinical applications, including cancer diagnosis, stem cell research, and immunology. Manual counting is labor-intensive and error-prone, motivating automation through deep learning techniques. However, training reliable deep learning models requires large amounts of high-quality annotated data, which is difficult and time-consuming to produce manually. Consequently, existing cell-counting datasets are often limited, frequently containing fewer than $500$ images. In this work, we introduce a large-scale annotated dataset comprising $3{,}023$ images from immunocytochemistry experiments related to cellular differentiation, containing over $430{,}000$ manually annotated cell locations. The dataset presents significant challenges: high cell density, overlapping and morphologically diverse cells, a long-tailed distribution of cell count per image, and variation in staining protocols. We benchmark three categories of existing methods: regression-based, crowd-counting, and cell-counting techniques on a test set with cell counts ranging from $10$ to $2{,}126$ cells per image. We also evaluate how the Segment Anything Model (SAM) can be adapted for microscopy cell counting using only dot-annotated datasets. As a case study, we implement a density-map-based adaptation of SAM (SAM-Counter) and report a mean absolute error (MAE) of $22.12$, which outperforms existing approaches (second-best MAE of $27.46$). Our results underscore the value of the dataset and the benchmarking framework for driving progress in automated cell counting and provide a robust foundation for future research and development.
\end{abstract}


\section{Introduction}
Cell counting is a fundamental task in biomedical research, essential for quantifying cell populations, studying cellular dynamics, and investigating complex biological processes. Accurate and scalable cell counting is critical for disease diagnosis, monitoring disease progression~\cite{orth_microscopy_2017, blumenreich1990white}, helping identify biomarkers~\cite{drost2018organoids, das_electrical_2017}, and evaluating treatment responses~\cite{polley_international_2013}. It supports drug discovery by screening therapeutic candidates and analyzing drug efficacy, and is indispensable in regenerative medicine to assess treatment outcomes.

In neural regeneration research, precise quantification of neural stem cells, neurons, and glial cells across developmental stages and injury models helps to uncover mechanisms underlying neurogenesis and repair. These insights have direct implications for the treatment of neurological disorders such as stroke, spinal cord injury, and neurodegenerative diseases. Cell counting is critical for stem cell therapy research, which attempts to use adult stem cells from patients~\cite{doi:10.1089/ten.2006.12.1451, BROHLIN200941, dezawa2001sciatic, ladak2011differentiation, sharma2016oriented} to treat their peripheral nerve injuries.

Traditionally, cell counting has relied heavily on manual annotation, which is labor-intensive, time-consuming, and challenging to scale for large datasets or images with dense cells (i.e., high-density cell images). Three paradigms of automated methods have emerged: the \emph{detection-based}, \emph{regression-based}, and \emph{density map estimation} methods. 

Detection-based methods localize individual cells using bounding boxes or segmentation masks, typically via convolutional neural networks or transformer-based detectors. While offering high interpretability and enabling instance-level analysis, these methods often struggle with dense regions due to overlapping cells and non-maximal suppression errors. Regression-based methods directly predict a single scalar count, reducing annotation effort but sacrificing the ability to localize cells and making annotation difficult to trace. Density map estimation (DME) approaches predict a 2D density map where the sum of all pixel values corresponds to the total count. DME was widely used for crowd counting~\cite{li_csrnet_2018, zhang_single-image_2016}, of which few were adopted for cell counting. This approach provides spatial context, robustness to occlusion, and the ability to train from limited annotations (dot annotations) compared to detection-based methods.

Despite significant advancements, automated cell counting remains a challenge. Existing automated methods are hindered by the scarcity of large, diverse, and consistently annotated datasets. Public benchmarks with limited annotations, such as \cite{lempitsky_learning_2010,kainz_you_2015,marsden_people_2018,paul_cohen_count-ception_2017,mohammed_idcia_2023}, have driven the field forward but are limited in scale, density variation, and biological diversity. 
\begin{table*}[!h]
\centering
\caption{Microscopy cell counting datasets with dot annotations}
\label{tab:datasets}
\footnotesize
\begin{tabular}{p{2.8cm}p{2.0cm}p{1.3cm}p{1cm}p{2.0cm}p{3.0cm}}
\hline
\textbf{Datasets}  & \textbf{Cell Types} & \textbf{\# Images} & \textbf{\#Cells} & \textbf{Mean CPI $\pm$ std} & \textbf{Image Size} \\ \hline
\rowcolor{syntheticcolor}
VGG~\cite{lempitsky_learning_2010}  & Bacterial & 200 & 35,192 & 176 ± 61 & 256x256 \\
\rowcolor{brightfieldcolor}
MBM~\cite{kainz_you_2015}  & Bone marrow & 44 & 5,553 & 126 ± 33 & 600x600\\
\rowcolor{brightfieldcolor}
ADI~\cite{paul_cohen_count-ception_2017}   & Adipose tissue & 200 & 31,017 & 165 ± 44  & 150x150\\
\rowcolor{brightfieldcolor}
DCC~\cite{marsden_people_2018}   & Various types & 176 & 5,906 & 34 ± 22 & 306×322 to 798×788\\
\rowcolor{fluorescencecolor}
IDCIA~\cite{mohammed_idcia_2023}  & AHPC & 262 & 22,155 & 84 ± 104 & 800x600\\
\hline
\rowcolor{fluorescencecolor}
CellFMCounter (ours)  &  AHPC \& rpc & 3,023 & 431,321 & 143 ± 316 & 600×447 to 1600×1200\\
\hline
\multicolumn{6}{l}{\footnotesize\textit{\textbf{AHPC}: Adult Hippocampal Progenitor Cells. \textbf{rpc}: Murine Retinal Progenitor Cells. \textbf{CPI}: \#Cells per image}} \\
\multicolumn{6}{l}{\footnotesize \textcolor{syntheticcolor}{\rule{0.8em}{0.8em}} Synthetic \hspace{0.1em} \textcolor{brightfieldcolor}{\rule{0.8em}{0.8em}} Brightfield \hspace{0.1em} \textcolor{fluorescencecolor}{\rule{0.8em}{0.8em}} Fluorescence} \\
\end{tabular}
\end{table*}
{\bf Challenges for fluorescence microscopy cell counting} include extreme variability in cell densities (number of cells per image), ranging from fewer than ten to thousands per image, cell type, cell development stage, staining strategies, image magnification, and lack of clear object contrast as seen in crowd counting.

\begin{figure*}[htbp]
\centering
\includegraphics[width=\textwidth]{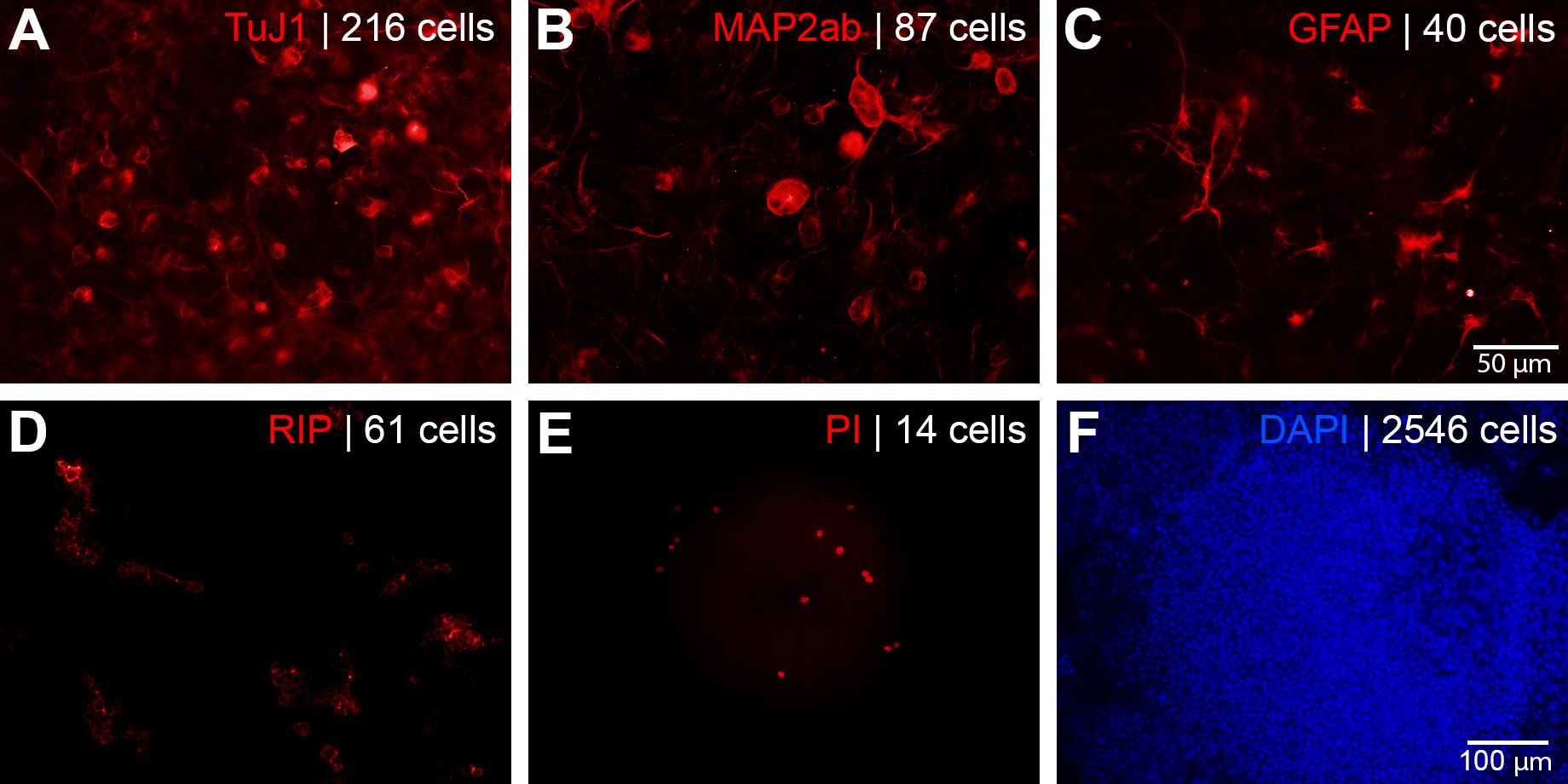}
% \caption{\textbf{Cell-type diversity and morphological variation across immunolabeled samples.} Cells were immunolabeled with antibodies for markers of specific cell types, including immature neurons (TuJ1), maturing neurons (MAP2ab), oligodendrocytes (RIP), astrocytes (GFAP), proliferating cells (Ki67), and dead cells (PI), along with nuclear staining (DAPI). Images show drastic differences in both cell morphology and density, ranging from sparse (e.g., PI, 14 cells) to dense populations (e.g., DAPI, 2546 cells). Scale bar = $\mu m$. \textit{Actual images in grayscale were pseudo-colored to enhance visualization.}}
\caption{\textbf{Cell-type diversity and morphological variation across immunolabeled samples.} Cells were immunolabeled with antibodies for markers of specific cell types, including immature neurons (TuJ1; A), maturing neurons (MAP2ab; B), astrocytes (GFAP; C), and oligodendrocytes (RIP; D). Cells were also stained with a cell viability dye to mark for dead cells (PI; E) as well as a nuclear stain (DAPI; F). Images show drastic differences in both cell morphology and density, ranging from sparse cells (e.g., PI, 14 cells) to dense fields (e.g., DAPI, 2546 cells). Scale bar = 50 $\mu m$ for 40x image fields (Row 1) or 100 $\mu m$ for 20x image fields (Row 2). Images were pseudo-colored for visualization.}
\label{fig:background}
\end{figure*}

\noindent\textbf{Contributions.}
To overcome these limitations, we introduce \textbf{CellFMCount}, a large-scale, annotated fluorescence microscopy dataset explicitly designed to facilitate robust and generalizable cell counting. CellFMCount contains 3,023 images from diverse biological contexts, imaging protocols, and staining conditions, featuring over 430,000 annotated cells. Each cell is labeled with a dot near its visual center, providing scalable supervision without requiring extensive segmentation annotations.
Beyond serving as a cell-counting benchmark, CellFMCount supports numerous downstream tasks. Its variability enables exploration of effective data augmentation and domain adaptation. Dot annotations facilitate weak supervision approaches, such as pseudo-labeling for detection-based cell counting. Additionally, the extensive scale and diversity of the dataset are suitable for investigating sample-efficient learning paradigms such as active learning.

We also present an adaptation of the Segment Anything Model (SAM)~\cite{kirillov2023segany}, SAM-Counter, for density-map estimation (DME)–based cell counting. Our implementation pairs SAM’s pretrained encoder with a lightweight density estimation head to generate density maps, showing competitive accuracy without requiring full segmentation masks.

In summary, we make the following contributions.
\begin{itemize}
\item Introduce \textbf{CellFMCount}, a large-scale, high-diversity, annotated fluorescence microscopy dataset for training and evaluating cell counting models under highly varying number of cells per image and appearance variation.
\item Demonstrate how SAM can be repurposed for DME-based counting through our SAM-Counter, highlighting the feasibility of leveraging foundation-model features for microscopy analysis.
\item Benchmark \textbf{thirteen cell-counting methods} with dot annotations as training ground truth. The methods include regression-based and density-based techniques, leading models adapted from the crowd-counting literature, and our SAM-Counter. Our evaluations provide unified and extensive baselines reflective of real-world complexity, setting a foundation for future research in automated microscopy-based cell counting.
\end{itemize}
% Upon acceptance of the paper, we will release all data, annotations, code, and pretrained models to promote transparency, reproducibility, and progress in automated microscopy analysis.

All data, annotations, code, and pretrained models have been released to promote transparency, reproducibility, and progress in automated microscopy analysis. The CellFMCount dataset is available at \url{https://doi.org/10.5281/zenodo.17088532}, and the code repository (including training scripts and pretrained SAM-Counter models) is available at \url{https://github.com/NRT-D4/CellFMCount}.

\section{Related Work}
\label{related_work}
This section describes existing datasets and methods for cell counting.
\subsection{Cell Datasets}
Many datasets were developed to support research in automatic cell counting and analysis, each highlighting different imaging techniques, biological samples, and annotation strategies. 

\iffalse
These resources play a crucial role in evaluating and comparing new algorithms, especially as methods aim to handle diverse conditions seen in practice.
\fi

\textbf{Cell datasets with segmentation masks or bounding boxes:} The ACCT dataset includes fluorescent images of neurons and microglia and corresponding segmentation masks for detection-based counting~\cite{kataras2023acct}. Similarly, the CIDACC dataset 
%focuses on microalgae and 
captures the particular difficulties in counting dense and irregularly shaped Chlorella vulgaris cells~\cite{pistolas2024cidacc}. The Nasal Cytology Dataset (NCD) addresses clinical diagnosis by including nasal cell images with annotated bounding boxes, accounting for rare cell types and uneven distributions~\cite{ncd_dataset_2024}.

\textbf{Cell datasets with dot annotations:} Table~\ref{tab:datasets} lists existing datasets in this category. {\em \textbf{Synthetic Datasets:}} One of the earlier and widely used datasets is the VGG Synthetic Fluorescence Microscopy Dataset~\cite{lempitsky_learning_2010}, which contains computer-generated images that mimic the look of real fluorescence microscopy images.
{\em \textbf{Real datasets:}} They better reflect the challenges found in actual biological samples. The Modified Bone Marrow (MBM)~\cite{kainz_you_2015} and Adipose Tissue (ADI)~\cite{paul_cohen_count-ception_2017} datasets include images of blood and fat tissue, respectively. These datasets highlight the variability in cell shape and distribution that occurs naturally across different tissue types.
The Dublin Cell Counting (DCC) dataset features a variety of cell images captured under different settings for evaluation of how well models generalize to new data~\cite{marsden_people_2018}. The IDCIA dataset has annotated fluorescence microscopy images of cells labeled with different antibody markers, which can significantly alter their appearance~\cite{mohammed_idcia_2023}.

Our CellFMCount dataset addresses the limitations of existing cell datasets in terms of dataset sizes, significant variability in cell densities, cell type, antibody and cell staining strategies, and image magnification.

\subsection{Methods for Cell Counting} 
Early techniques combined intensity thresholding, edge detection, and watershed-like algorithms to separate touching objects, then counted connected components~\cite{buggenthin2013automatic}. Although these methods are fast and annotation-free, they struggle when cell shape and contrast vary widely.
Tools such as CellProfiler~\cite{stirling2021cellprofiler} have extended traditional pipelines by integrating segmentation models with user-guided parameter tuning. However, they require users to specify object properties or provide annotated masks for model training or refinement, limiting scalability and automation, especially in heterogeneous or dense datasets.

Further developments involved methods focused on learning mappings from local image features to a density map. Lempitsky and Zisserman \cite{lempitsky_learning_2010} pioneered this line of research by employing linear regression with dense SIFT features to predict density maps. Subsequent work replaced linear regression with regression forests to enhance density estimation accuracy. Arteta et al.~\cite{ARTETA20163} extended this pipeline by introducing a local feature vocabulary and ridge regression in an interactive framework.

Recent methods shift toward deep neural networks. Xie et al.\cite{xie_microscopy_2018} applied fully convolutional regression networks (FCRN) with Gaussian-filtered outputs, while Count-ception~\cite{paul_cohen_count-ception_2017} used sliding-window aggregation over receptive fields to construct density maps, improving accuracy but risking overfitting in background regions. SAUNet~\cite{guo_sau-net_2019} enhanced U-Net with self-attention and modified batch normalization for small datasets, and Xue et al.\cite{xue_cell_2016} proposed partitioning images into sub-regions processed by separate networks. Wang et al.\cite{he_deeply-supervised_2021} introduced a concatenated fully convolutional regression network with auxiliary (CFCRN+Aux) supervision to improve intermediate feature learning and generalization across datasets.

A zero-shot SAM-based approach for cell counting was introduced \cite{fanijo2025idcc}, but is less accurate compared to the convolution neural network models trained on the IDCIA dataset. These efforts demonstrate SAM's versatility in addressing microscopy challenges, from complex morphologies to scalable analysis.

\section{Proposed Dataset: Data Curation}
We curated the images from stem cell differentiation experiments using mouse-derived Retinal Progenitor Cells (rpcs) and rat-derived Adult Hippocampal Progenitor Cells (AHPCs) to investigate the effects of different stimuli on neural progenitor cell differentiation.
Differentiating stem cells into mature and functional cells requires carefully designed experiments. A typical cell differentiation experiment might involve comparing the effect of different growth factor concentrations on cell differentiation of the cell population during a seven-day period. 

\begin{figure}[h]
\centering
\includegraphics[width=\linewidth]{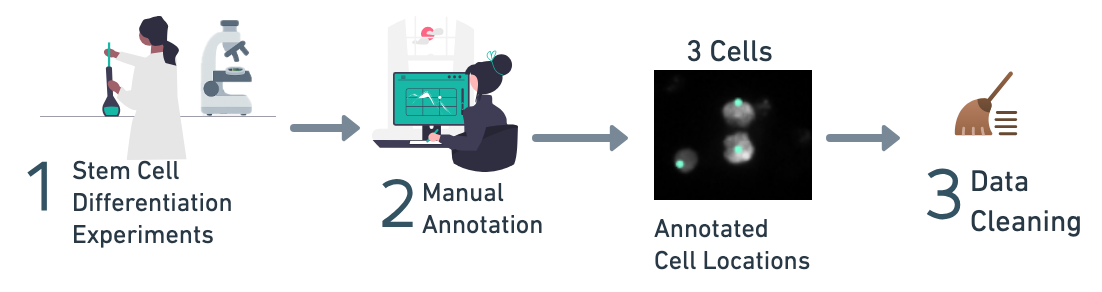}
\caption{CellFMCount dataset creation pipeline.}
\label{fig:process}
\end{figure}

Fig.~\ref{fig:process} shows the CellFMCount dataset creation pipeline from stem cell differentiation experiments, manual annotation of cell images, to data cleaning, producing high-quality annotated cell locations.

{\bf Step 1: Stem Cell Differentiation Experiments}

The biological experiments followed a typical process for stem cell differentiation. Once a starting growth factor concentration has been identified, additional concentrations above and below the literature-based determination were used. A control would be incorporated where no growth factor was added. When planning an experiment, duplicate samples are included just in case a mishap occurs and a sample is destroyed (4 concentrations of the growth factor $\times$ 2 samples/condition = 8 samples $\times$ number of antibodies to be screened (5 antibodies) = 40 total samples).

At the end of the seven days, these cell cultures were fixed and immunolabeled with a panel of cell-type-specific antibody markers (see Table~\ref{tab:marker-bio}), followed by a fluorescent secondary antibody, such as Cy3 (Cyanine 3) or AF488 (Alexa Fluor 488) conjugated secondary antibody, to visualize the immunolabeled cells. Cells can also be incubated in propidium iodide (PI), a fluorescent cell viability dye, to determine the cell viability of the cells at the conclusion of the experiment. Some experiments used co-labeling with two primary antibodies; however, during imaging, each fluorescent channel displayed only one marker at a time. Since images were not overlaid prior to analysis, markers were quantified separately.
All samples were also counterstained with a nuclear stain such as DAPI to facilitate counting the total cell population. All samples were imaged under fluorescence microscopy using a 20x or 40x objective. The above process is also known as immunocytochemistry (ICC). 

Ten image regions (fields) were selected systematically, and 2-3  images were captured (antibody immunolabeling and DAPI) for each image field per biological condition. The following counts would be made in each field: \textit{the total number of cells (DAPI-stained nuclei) and the number of cells expressing the primary antibody of interest}, Ki67, TuJ1, MAP2ab, RIP, or GFAP. This data would be used to calculate the percentage of cells immunolabeled with each respective antibody marker. For each antibody marker and each condition, we typically gathered 20 images (10 for DAPI-stained images and 10 for antibody-labeled image fields) 
for each condition. Students supervised by the biologist authors performed experiments and collected images. They used either an inverted Leica fluorescent microscope (Leica DMI4000B) equipped with standard epiflorescence illumination and a Leica DFC310 FX  digital camera or an upright Leica fluorescent microscope (Leica DM5000B) equipped with standard epifluorescence illumination and a Q Imaging Retiga 2000R (Q Imaging) digital camera. 

\begin{table}[h]
\centering
\caption{Summary of nuclear stains and immunolabeling markers with their biological relevance.}
\label{tab:marker-bio}
\begin{tabular}{ll}
\toprule
\textbf{Marker} & \textbf{Biological Target / Description} \\
\midrule
\textbf{DAPI}             & Universal nuclear stain (labels DNA in all cells) \\
\textbf{PI}               & Propidium Iodide; labels dead or permeable nuclei \\
\textbf{RIP}              & Oligodendrocyte lineage marker\\
\textbf{GFAP}             & Astrocyte marker \\
\textbf{TuJ1}             & Neuronal marker for immature neurons \\
\textbf{MAP2ab}           & Neuronal marker for maturing neurons \\
\textbf{Ki67}           & Proliferating Cells \\
% \textbf{Ki67 + TuJ1}      & Actively proliferating neurons \\
% \textbf{MAP2ab + RIP}     & Co-labeling of dendritic processes and oligodendrocytes \\
% \textbf{GFAP + Ki67}      & Proliferating astrocytes \\
% \textbf{Ki67 + RIP}       & Proliferating oligodendrocyte lineage cells \\
\bottomrule
\end{tabular}
\end{table}

{\bf Step 2: Data Annotation}

To perform manual annotation, all images were imported into ImageJ (FIJI distribution) \cite{schneider_nih_2012}.  Graduate and undergraduate students trained to evaluate antibody labeled images performed the initial annotation on their assigned sets of images. The students used ImageJ CellCounter plugin to place a single dot on each nucleus, marking each cell of interest by clicking directly on the image. To ensure high annotation fidelity, every undergraduate annotation set was reviewed and corrected as needed by a graduate student with experience in cell analysis. The final dot coordinates and associated metadata were exported from CellCounter in XML format, preserving annotation structure. 

{\bf Step 3: Data Cleaning}

This process ensured that all annotations were complete, accurate, and machine learning-ready. First, we verified that each XML file exported from the ImageJ CellCounter plugin matched its corresponding microscopy image. Any XML files without a corresponding image or vice versa were removed to maintain data integrity. 
The duplicate image annotation pairs were then removed to prevent data leakage and biased performance estimates during model development and evaluation. All image and annotation filenames were standardized to numerical identifiers to simplify file referencing and ensure compatibility with automated processing pipelines. A separate metadata CSV file was created to maintain traceability, containing the original filenames, associated antibody or cell staining markers, imaging magnification, and other relevant experimental details for each sample.

The original dot annotations in XML format were converted into machine learning-friendly CSV files. Each CSV file contains the $X$ and $Y$ coordinates of all annotated cells in the corresponding image. This format facilitates straightforward integration into different machine learning pipelines used in counting tasks.

Following this cleaning and standardization process, the final CellFMCount dataset comprises 3,023 fluorescence microscopy images containing over 430,000 manually annotated cell locations.

\section{Dataset Description}

\subsection{Staining Modalities and Cell Count Statistics}

% The CellFMCount dataset includes images stained using three imaging channels: DAPI, Cy3, and AF488. These fluorophores highlight different cellular components, contributing to variability in image appearance and cell counts. Of the 3,023 total images, approximately 45\% include DAPI, 47\% Cy3, and the remainder AF488 (Table~\ref{tab:stain-summary}). DAPI is used as a nuclear counterstain alongside other markers and is essential for determining the total number of cells in each image, as it labels all nuclei regardless of the specific marker being targeted. 
% The differences in average and maximum cell counts reflect the biological specificity of each label. DAPI binds to DNA and labels all nuclei, making DAPI-stained images denser. In contrast, secondary antibodies conjugated with either Cy3 and AF488 indicate cellular structures or subpopulations, resulting in lower and sparse cell densities. The distributions of cell counts for DAPI, Cy3, and AF488 are highly skewed, with most images containing relatively few cells. 

The CellFMCount dataset includes images stained with three channels: DAPI, Cy3, and AF488. Of the 3,023 images, approximately 45\% include DAPI, 47\% Cy3, and the rest AF488 (Table~\ref{tab:stain-summary}). DAPI is routinely used together with other markers as a universal nuclear counterstain, which explains its high representation in the dataset and makes it essential for determining the total cell population. DAPI-stained images are therefore denser, while Cy3 and AF488 along with other markers highlight specific structures or subpopulations and produce sparser images. Across all channels, cell counts follow a long-tailed distribution with most images containing relatively few cells.

\begin{table}[h]
\centering
\caption{Summary statistics of cell counts across imaging channels. CPI: \#Cells per image.}
\label{tab:stain-summary}
\begin{tabular}{lcccc}
\toprule
\textbf{Stain} & \textbf{\#Images} & \textbf{\#Cells} & \textbf{Mean CPI$\pm$ std} & \textbf{Min / Max CPI}\\
\midrule
DAPI  & 1,373 & 386,702 & 281.6 $\pm$ 426.9 & 4~/~2,546 \\
Cy3   & 1,428 & 42,797 & 30.0  $\pm$ 51.2  & 0~/~383 \\
AF488 & 222   & 1,822 & 8.2   $\pm$ 16.6  & 0~/~128 \\

\bottomrule
\hline
\end{tabular}
\end{table}

\subsection{Marker Diversity and Labeling Strategies}

CellFMCount includes six immunocytochemistry markers that identify different biological structures or cellular states involved in neural differentiation. Depending on the experimental design, these markers are applied individually or in various combinations. They are used to label nuclei, neuronal structures, glial cells, or indicators of proliferation. This supports a detailed characterization of cellular heterogeneity and phenotypic diversity during the differentiation process.

Table~\ref{tab:marker-summary-table} shows descriptive statistics of the dataset by antibody marker type. \textbf{Ki67 + TuJ1} denotes that both Ki67 and TuJ1 were used (colabeling). Ki67 marks actively proliferating cells, while TuJ1 highlights immature neurons, enabling identification of proliferating neuronal precursors. Similarly, combinations like \textbf{MAP2ab + RIP} or \textbf{GFAP + Ki67} allow assessment of phenotypically distinct or transitioning cell states.

\begin{table}[h]
\centering
\caption{Cell count statistics for each marker type. Markers with low average counts and high standard deviation may present detection challenges. CPI: \#Cells per image}
\label{tab:marker-summary-table}
\resizebox{0.9\linewidth}{!}{%
\begin{tabular}{lcccc}
\toprule
\textbf{Marker} & \textbf{\#Images} & \textbf{Mean CPI} $\pm$ \textbf{std} & \textbf{Median CPI} & \textbf{Min} / \textbf{Max CPI} \\
\midrule
DAPI             & 1,373 & 281.6 $\pm$ 426.9 & 110   & 4 / 2,546 \\
PI               & 552   & 13.4  $\pm$ 36.1  & 0    & 0 /  12 \\
Ki67 + TuJ1      & 295   & 37.0  $\pm$ 59.0  & 10   & 0  / 383 \\
MAP2ab           & 242   & 20.9  $\pm$ 23.03  & 11    & 0  / 128 \\
GFAP + Ki67      & 178   & 1.3   $\pm$ 2.39   & 0    & 0  / 12 \\
RIP              & 120   & 18.1  $\pm$ 13.24  & 15    & 2  / 67 \\
TuJ1             & 119   & 74.1  $\pm$ 49.33 & 71   & 1  / 216 \\
Ki67 + RIP       & 67    & 129.9 $\pm$ 79.45 & 131   & 1  / 318 \\
GFAP             & 57    & 10.4  $\pm$ 19.12  & 1    & 0  / 84 \\
MAP2ab + RIP     & 20    & 39.2  $\pm$ 24.35  & 35.5   & 6  / 91 \\
\bottomrule
\end{tabular}
}
\end{table}

\textbf{Dataset Challenges.}
The wide range in image counts and average cell densities across markers introduces significant class imbalance and sparsity. For instance, DAPI-stained images are dense and abundant, while markers like GFAP, MAP2ab, TuJ1, and RIP occur in fewer than 300 images and often exhibit very low cell counts. This imbalance introduces challenges for model generalization, particularly in the context of density map-based cell counting. Markers corresponding to rare cellular populations yield highly sparse density maps, making learning more difficult and prone to overfitting. 
Fig.~\ref{fig:background} shows cell images with varying morphologies across markers. DAPI highlights densely packed nuclei, while MAP2ab reveals elongated structures. These differences introduce spatial complexity that challenges standard feature extraction methods.

We restrict our benchmarking to the DAPI-stained subset, which offers both high cell density and consistent availability across conditions, providing a stable and representative basis for model assessment.

\subsection{DAPI-stained Subset Characterization}

 The DAPI-stained image subset includes 1,373 grayscale fluorescence microscopy images with 386,702 annotated cell locations. Table~\ref{tab:marker-summary-table} shows highly variable cell count per image (CPI), with a long-tailed distribution. While the average is 281.6 cells per image (std: 426.9), over 25\% of images contain fewer than 38 cells, and some images contain over 2,500 cells. This reflects substantial biological heterogeneity, differences in experimental conditions (e.g., growth conditions), and variation in the field of view.

Table~\ref{tab:celltype-summary} shows statistics of the DAPI-stained subset by cell types. Two neural progenitor cell types: adult rat hippocampal progenitor cells (AHPC) and murine retinal progenitor cells (RPC), were imaged at 20$\times$ and 40$\times$ magnification, respectively. Notably, RPC images exhibit higher average cell counts and lower variance in density despite having smaller fields of view.

\begin{table}[h]
\centering
\caption{Descriptive statistics for DAPI-stained AHPC and RPC subsets. CPI: \#Cells per image.}
\label{tab:celltype-summary}
\resizebox{0.9\linewidth}{!}{%
\begin{tabular}{lcccc}
\toprule
\textbf{Cell type} & \textbf{\#Images} &  \textbf{Mean CPI$ \pm$ std} & \textbf{Median} & \textbf{Min~/~Max CPI}\\
\midrule
\textbf{AHPC (20$\times$)}  & 1,214 &  279.3 $\pm$ 451.7  & 93 & 4~/~2,546 \\
\textbf{RPC (40$\times$)}   & 159 &  299.7 $\pm$ 125.6   & 258 &  75~/~686 \\

\bottomrule
\hline
\end{tabular}
}
\end{table}

\section{Model Benchmarking}
\label{benchmark}

% We present a comprehensive evaluation of  twelve established methods including regression‑based, density‑map, and crowd‑counting approaches as well as a case study adaptation of the Segment Anything Model (SAM) for density‑map estimation. All experiments were on DAPI‑stained images from our CellFMCount dataset, which feature high cell densities, overlapping nuclei, and staining variability. By benchmarking across counts ranging from ten to over two thousand cells per image, we assess each method’s accuracy and robustness under challenging, real‑world conditions, illustrating the potential for significant reductions in manual counting effort.
We present a comprehensive evaluation of twelve established methods, including regression-based, density-map, and crowd-counting approaches, as well as a case study adaptation of the Segment Anything Model (SAM) for density-map estimation. For this study, we focused on DAPI-stained images from CellFMCount, as they are the most abundant, display consistent nuclear contrast, and represent a widely used staining protocol in microscopy. These images feature high cell densities, overlapping nuclei, and staining variability, making them a challenging yet representative benchmark. Future work will extend our analysis to other markers (e.g., Cy3, AF488) to study cross-channel generalization and model robustness in multimodal settings.

\subsection{Problem Formulation}

Given an input image $\mathbf{x} \in \mathbb{R}^{c \times H \times W}$ with $c$ channels and an image height and width of $H$ by $W$, the goal is to estimate the number of cells present in $\mathbf{x}$.

\paragraph{Regression-based Counting}
The objective is to learn a function $f_\theta: \mathbb{R}^{c \times H \times W} \rightarrow \mathbb{R}_{\geq 0}$ that maps an input image to a scalar prediction \( \hat{y} \) denoting the predicted cell count.
\begin{equation}
    \hat{y} = f_\theta(\mathbf{x})
\end{equation}
The model parameters \( \theta \) are optimized by minimizing a task-specific objective $\mathcal{L}_{\text{reg}}$ over the training dataset:
\begin{equation}
    \argmin_{\theta} \; \mathcal{L}_{\text{reg}}(f_\theta(\mathbf{x}), y),
\end{equation}
where \( y \in \mathbb{R}_{\geq 0} \) is the ground truth cell count for image \( \mathbf{x} \).

\paragraph{Density-based Counting}
Alternatively, counting can be formulated as estimating a density map \( D_\theta(\mathbf{x}) \in \mathbb{R}^{H_f \times W_f} \) and then summing the density values over the entire estimated density map.
\begin{equation}
    \hat{y} = \sum_{h=1}^{H_f} \sum_{w=1}^{W_f} D_{\theta}( \mathbf{x} )_{h,w},
    \label{eq:total_count}
\end{equation}
where \( H_f \) and \( W_f \) are the height and width of the output.
The density map ground truth \( D(\mathbf{x}) \) is typically constructed by placing normalized Gaussian kernels at annotated cell locations. The parameters \( \theta \) are learned by minimizing a density-specific objective $\mathcal{L}_{\text{dens}}$.
\begin{equation}
    \argmin_{\theta} \; \mathcal{L}_{\text{dens}}(D_\theta(\mathbf{x}), D(\mathbf{x})).
\end{equation}

Both formulations aim to infer accurate cell counts from microscopy images, with the density-based approach additionally offering localization information.

\subsection{Proposed SAM-Counter Model}

\begin{figure}[h]
\centering
\includegraphics[width=\linewidth]{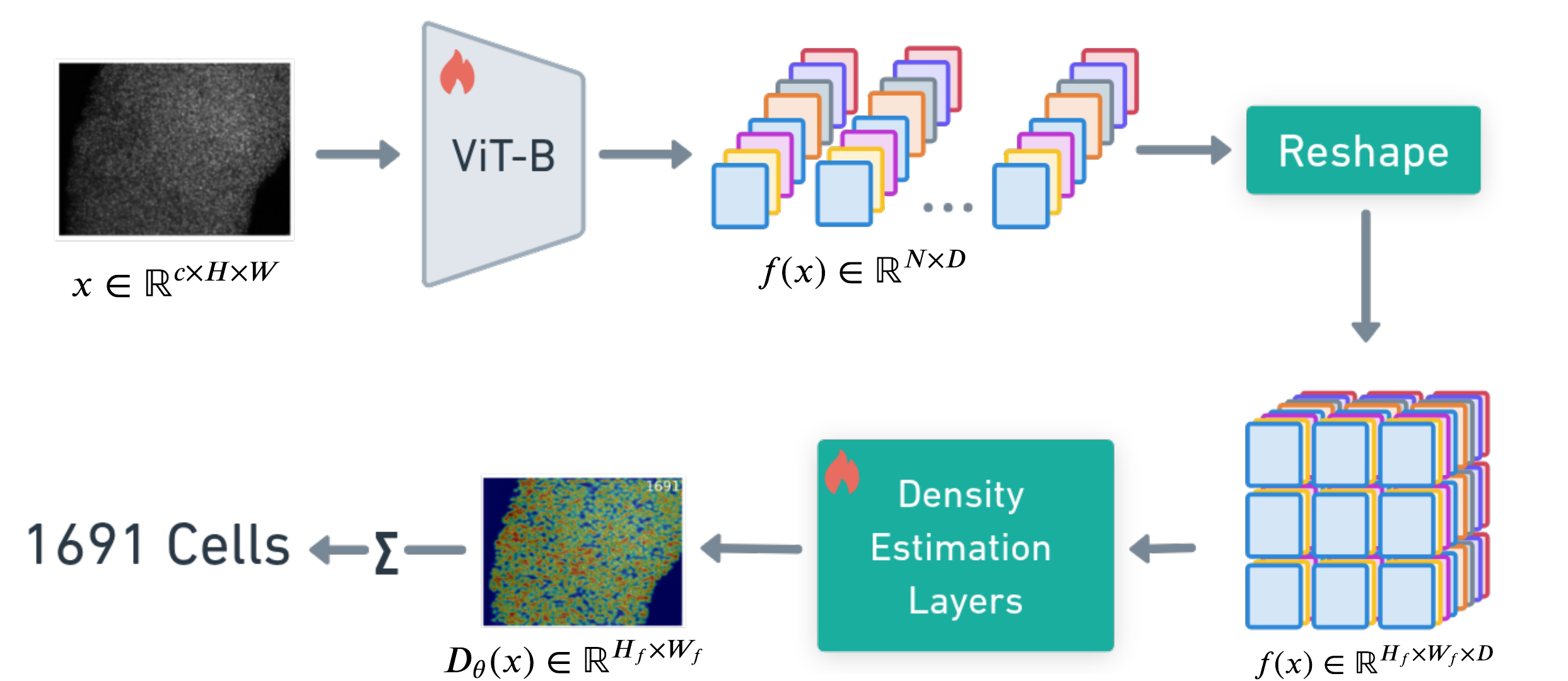}
\caption{
The architecture of SAM-Counter. The input image is encoded using a fine-tuned ViT encoder from SAM~\cite{kirillov2023segany}. The density estimation layers then estimate the density map, whose sum yields the cell count. See (~\ref{eq:feature_map}) for reshaping.
}
\label{fig:vit_pipeline}
\end{figure}

SAM‑Counter repurposes the pretrained Segment Anything Model (SAM) image encoder~\cite{kirillov2023segany} as the backbone of a density‑map estimation pipeline for cell counting. We remove SAM’s prompt encoders and mask decoder, retaining only the Vision Transformer (ViT) image encoder, which is then paired with lightweight convolutional layers to produce density maps. See Fig.~\ref{fig:vit_pipeline}.

SAM's encoder, originally trained on over one billion segmentation masks sourced from diverse natural images, provides spatial representations with rich semantic structure and strong cross‑domain generalization. To adapt these features to microscopy data, we reshape the ViT output and feed it to lightweight convolutional layers that produce density maps for cell counting. We then fine‑tune the entire encoder–decoder stack on our annotated cell images, enabling the model to capture domain‑specific characteristics such as variable cell morphology, staining variability, and imaging noise. The resulting model is highly accurate, leveraging SAM’s large‑scale pretraining to achieve robust performance even with limited training samples.

% Given an input image \( \mathbf{x} \in \mathbb{R}^{c \times H \times W} \), the SAM-encoder divides it into non-overlapping patches of size \( P \times P \). Each patch is flattened and projected into a \( D \)-dimensional embedding via a learned linear projection. These patch embeddings are passed through a Vision Transformer (ViT), producing a sequence of contextualized embeddings, which we reshape into a spatial feature map:
% \begin{equation}
% \mathbf{f}(\mathbf{x}) \in \mathbb{R}^{H_f \times W_f \times D}, \quad \text{where } H_f = \frac{H}{P},\; W_f = \frac{W}{P}.
% \label{eq:feature_map}
% \end{equation}

Given an input image \( \mathbf{x} \in \mathbb{R}^{c \times H \times W} \), the SAM-encoder divides it into non-overlapping patches of size \( P \times P \). Each patch is flattened and projected into a \( D \)-dimensional embedding, producing \( N=\tfrac{H}{P}\cdot\tfrac{W}{P} \) tokens. Passing these through a Vision Transformer (ViT) yields contextualized embeddings that we reshape into a spatial feature map:
\begin{equation}
\mathbf{f}(\mathbf{x}) \in \mathbb{R}^{H_f \times W_f \times D}, \quad \text{where } H_f = \tfrac{H}{P},\; W_f = \tfrac{W}{P}.
\label{eq:feature_map}
\end{equation}

The spatial feature map \( \mathbf{f}(\mathbf{x}) \) is processed by convolutional density estimation layers. These layers consist of sequential convolutional blocks, implemented as a series of 1×1 convolutional layers with ReLU activations. The final 1×1 convolution reduces the channel dimension to 1, yielding a single-channel density map. 

We selected the ViT-Base (ViT-B) architecture as the encoder to balance performance and computational efficiency. Preliminary experiments show that ViT-Large and ViT-Huge offer only marginal improvements in performance~\cite{kirillov2023segany}. In contrast, these larger models significantly increased memory usage, training time and inference latency.

We fine-tune the SAM encoder jointly with the density estimation head in an end-to-end manner using Mean Squared Error (MSE) loss. This allows the model to adapt to domain-specific features in microscopy data. Freezing the encoder led to poor performance in our experiments (see Section~\ref{sec:ablation}). 
% {\color{red} TODO: Why repeating these equations again here? Can you simply refer to the similar equations earlier.}

% \begin{equation}
% D_\theta(\mathbf{x}) \in \mathbb{R}^{H_f \times W_f}.
% \label{eq:density_map}
% \end{equation}

% The predicted cell count \( \hat{y} \in \mathbb{R}_{\geq 0} \) is computed by summing the values of the density map as shown in Eq.~\ref{eq:total_count}

\subsection{Evaluation Metrics}

We utilize the following metrics: Mean Absolute Error (MAE), Mean Squared Error (MSE), Root Mean Squared Error (RMSE), Mean Absolute Percentage Error (MAPE), and Acceptable Count Percentage (ACP) \cite{mohammed_idcia_2023}. See Table~\ref{tab:performance-metrics}.

Let \( y_i \in \mathbb{R}_{\geq 0} \) denote the ground truth cell count for the \( i \)-th image, \( \hat{y}_i \in \mathbb{R}_{\geq 0} \) the predicted count, and \( n \in \mathbb{N} \) the total number of test samples. The indicator function \( \llbracket \cdot \rrbracket \) returns 1 if the condition inside is true, and 0 otherwise.

\begin{table}[htbp]
\centering
\scriptsize
\setlength{\tabcolsep}{8pt} 
\caption{Common performance metrics}
\begin{tabular}{p{0.40\linewidth}p{0.40\linewidth}}
\midrule % Midlines
$\displaystyle
\text{MAE}
=\frac{1}{n}\sum_{i=1}^{n}\bigl|y_i - \hat y_i\bigr|
$
&
$\displaystyle
\text{RMSE}
=\sqrt{\frac{1}{n}\sum_{i=1}^{n}(y_i - \hat y_i)^2}
$ \\[2ex]
\midrule % Midlines
$\displaystyle
\text{MAPE}
=\frac{100\%}{n}\sum_{i=1}^{n}
\biggl|\frac{y_i - \hat y_i}{y_i}\biggr|
$
&
$\displaystyle
\text{MSE}
=\frac{1}{n}\sum_{i=1}^{n}(y_i - \hat y_i)^2
$ \\[2ex]
\midrule % Midlines
\multicolumn{2}{p{0.8\linewidth}}{
$\displaystyle
\text{ACP}
=\frac{100\%}{n}\sum_{i=1}^{n}
\llbracket\,\bigl|\hat y_i - y_i\bigr|\le0.05\,y_i\rrbracket
$
} \\[2ex]

\hline
\end{tabular}
\label{tab:performance-metrics}
\end{table}

All the metrics except ACP capture absolute and relative prediction errors. Lower values are desirable. 
ACP follows the domain experts' desire to accept the result only when the predicted count is within a practically acceptable error tolerance of 5\%. Higher ACP values are desirable. 

\subsection{Training and Testing Sets}
\begin{figure} [h]
  \centering
  \includegraphics[width=1\linewidth]{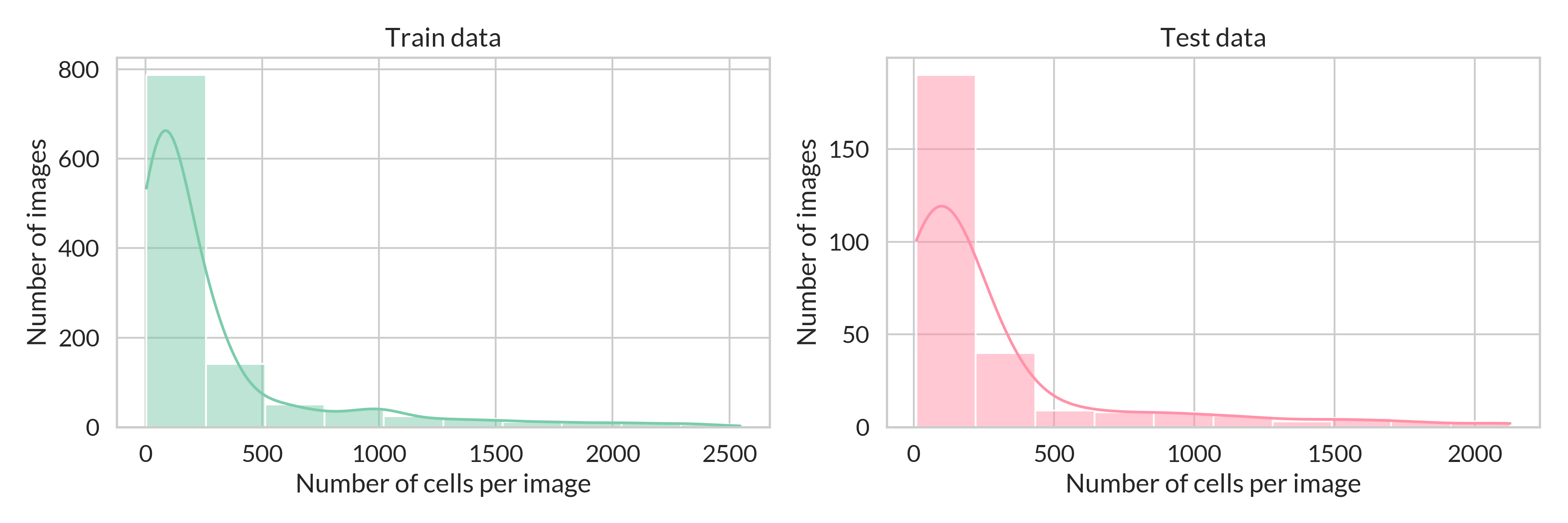}
  \caption{Distribution of cell counts per DAPI-stained images in training and testing sets.}
  \label{fig:train_test}
\end{figure}

All DAPI images were split into two non-overlapping sets: training (80\%) and testing (20\%) sets. To ensure representative training and testing sets despite the long-tailed distribution of cell counts, we employed a stratified splitting strategy. First, we discretized the cell count into 5 distinct bins using Jenk's natural breaks optimization \cite{jenks}. Stratification was then performed based on the combined categories of these cell count bins and the magnifications used, ensuring proportional representation in both the training and test sets. This approach aims to maintain similar cell count and magnification distributions across splits. As shown in Fig. \ref{fig:train_test}, both the training and test sets exhibit similar cell count distributions.

\subsection{Model Selection}
\textbf{Thirteen} distinct models were evaluated for cell counting, including CNN-based regression-based models and density estimation methods, and crowd-counting models. These models were chosen for their strong performance in cell counting and in other domains. The regression-based models employ a backbone feature extractor network followed by a fully connected layer to output a scalar value. Four different pre-trained backbones, namely VGG-16 \cite{vgg16_liu}, ResNet-50, ResNet-18 \cite{He2015DeepRecognition}, and EfficientNet B7 \cite{efficientnet_tan} were fine-tuned on our training dataset. The density map estimation models include four cell counting models \cite{guo_sau-net_2019,he_deeply-supervised_2021, paul_cohen_count-ception_2017, xie_microscopy_2018} and three crowd-counting models that were previously adopted for cell-counting tasks \cite{li_csrnet_2018, zhang_single-image_2016, lin2022boosting}. We used the original authors' implementations of the compared cell counting and crowd counting methods with a minor modification to use our dataset. We resized all images to 224x224 for faster training and because the pretrained models were trained on this image size. The ground truth density maps were generated by convolving each dot annotation with a normalized $5 \times 5$ Gaussian kernel. This kernel size offers a good balance between smoothing and preserving detail in areas with densely packed cells.
No augmentations were used unless the original implementation uses augmentation during training. The training of all the models involved searching for optimal values for batch size and learning rate based on a validation split. The model checkpoints with the lowest validation MAE were saved. Table~\ref{tab:model_hparams} shows the optimal hyperparameter values. A single round of SAM-Counter training on one NVIDIA A100 GPU took~36 hours.

% {\color{red} TODO: Need to discuss the hyperparameter values here. Reviewers can easily reject this article because of the lack of repeatability of the experiments. If you do not have space, you can indicate that the parameter values will be put in the shared code, and the models will be shared. Mention about the data card and model cards for sharing the data and models.}

\begin{table}[h]
    \caption{Optimal hyperparameter values based on validation MAE for each model under study}
    \label{tab:model_hparams}
    \centering
    \renewcommand{\arraystretch}{1.2}
    \resizebox{0.6\linewidth}{!}{%
    \begin{tabular}{l c c}
        \toprule
        \textbf{Model} & \textbf{Batch size} & \textbf{Learning rate} \\
        \midrule
        ResNet-18~\cite{He2015DeepRecognition}            & 8    & $4.92\times10^{-4}$ \\
        ResNet-50~\cite{He2015DeepRecognition}            & 16   & $7.8\times10^{-3}$  \\
        VGG-16~\cite{vgg16_liu}                 & 8    & $2.57\times10^{-4}$ \\
        EfficientNet\,B7~\cite{efficientnet_tan}       & 16   & $8.667\times10^{-4}$ \\
        MCNN~\cite{zhang_single-image_2016}                   & 32   & $3.79\times10^{-4}$ \\
        CSRNet~\cite{li_csrnet_2018}                 & 32   & $3.31\times10^{-3}$ \\
        MAN~\cite{lin2022boosting} & 4 & $1.0\times10-5$ \\
        SAUNet~\cite{guo_sau-net_2019}                 & 75   & $1.0\times10^{-3}$ \\
        C-FCRN+AUX~\cite{he_deeply-supervised_2021}             & 100  & $1.0\times10^{-3}$ \\
        Count-ception~\cite{paul_cohen_count-ception_2017}           & 32   & $5.166\times10^{-3}$ \\
        FCRN-A~\cite{xie_microscopy_2018}                 & 16   & $9.708\times10^{-3}$ \\
        SAM-Counter\,(Ours)    & 8    & $1.0\times10^{-6}$ \\
        \bottomrule
    \end{tabular}%
    }
\end{table}

% All models except MAN and IDCC SAM were implemented
% Models that use augmentations
% SAUNet - Flipping (Horizontal and Vertical)
% C-FCRN - Rotation, Random Crop, Flip, and Pad if needed

\section{Results}
Table~\ref{tab:model_performance} reports the performance of the thirteen compared models on the test dataset. In all performance tables, the best performance for each method category is underlined. The best overall performances are bold.

\begin{table}[h]
\caption{Performance comparison of different models on the test data }
    \label{tab:model_performance}
    \centering
    \renewcommand{\arraystretch}{1.2}
    \resizebox{0.9\linewidth}{!}{%
    \begin{tabular}{l c c c c c}
        \toprule
        \textbf{Model}  & \textbf{MAE} $\downarrow$ & \textbf{MSE} $\downarrow$ & \textbf{RMSE} $\downarrow$ & \textbf{MAPE} $\downarrow$ & \textbf{ACP} $\uparrow$ \\
        \midrule
        \rowcolor{Gray} \multicolumn{6}{l}{\textbf{Regression-based}} \\
ResNet-18 \cite{He2015DeepRecognition}      & 63.49  & 16651.17 & 129.04 & 30\% & 13.82\%\\
ResNet-50 \cite{He2015DeepRecognition}      & 61.73  & 14058.55 & 118.57 & 33\% & 11.64\%\\
VGG-16  \cite{vgg16_liu}       & \underline{38.15}  & \underline{7131.87}  & \underline{84.45}  & \underline{17\%} & \underline{24.73\%}\\
EfficientNet B7 \cite{efficientnet_tan}   & 41.76  & 8189.01  & 90.49  & \underline{17\%} & 19.27\%\\
\midrule
\rowcolor{Gray} \multicolumn{6}{l}{\textbf{Adopted Density Map Estimation from Crowd Counting}} \\
MCNN   \cite{zhang_single-image_2016}       & 42.31  & 9943.65  & 99.72  & 18\% & 25.09\%\\
CSRNet  \cite{li_csrnet_2018}     & \underline{27.46}  & \underline{4576.63}  & \underline{67.65}  & \underline{\textbf{9\%}} & 38.55\%\\
MAN \cite{lin2022boosting}          & 27.68  & 5014.03  & 70.81  & \underline{\textbf{9\%}} & \underline{42.55\%}\\
\midrule
\rowcolor{Gray} \multicolumn{6}{l}{\textbf{Density Map Estimation for Cell Counting}} \\
SAUNet   \cite{guo_sau-net_2019}     & 42.02  & 7866.28  & 88.69  & 24\% & 22.91\%\\
C-FCRN+AUX \cite{he_deeply-supervised_2021}    & 37.45  & 15133.47 & 123.02 & 14\% & 33.82\%\\
Count-ception \cite{paul_cohen_count-ception_2017}  & 34.79  & 7770.48  & 88.15  & 13\% & \underline{34.91\%}\\
FCRN-A \cite{xie_microscopy_2018}       & \underline{31.96}  & \underline{6616.88}  & \underline{81.34}  & \underline{12\%} & 32.36\%\\
        \midrule
        \rowcolor{Gray} \multicolumn{6}{l}{\textbf{SAM-based Cell Counting}} \\
        IDCC-SAM (zero shot)~\cite{fanijo2025idcc}   & 211.72  & 217437.74  & 466.3  & 54\% & 2.55\% \\
        \textbf{SAM-Counter (Ours)}   & \textbf{\underline{22.12}}  & \textbf{\underline{2470.71}}  & \textbf{\underline{49.71}}  & \underline{11\%} & \textbf{\underline{48.73\%}} \\
        \bottomrule
    \end{tabular}
    }
\end{table}

Across all evaluated categories, the proposed \textbf{SAM-Counter} achieves the strongest overall performance. It attains the lowest MAE (22.12), MSE (2470.71), and RMSE (49.71), alongside the highest ACP (48.73\%), indicating both accuracy and reliability in count prediction. CSRNet and MAN stand out among the baselines, suggesting robustness to input variability. They both tie for the minimal MAPE (9\%) with MAN offering stronger ACP (42.55\%).

Regression-based models such as ResNet-18 and ResNet-50 underperform across all metrics, with notably higher MAE and MSE values. In contrast, classical crowd-counting models (e.g., CSRNet, MAN) and domain-specific architectures (e.g., Count-ception, FCRN-A) show consistently better alignment with the spatial distribution of cellular features. The poor performance of IDCC-SAM (MAE of 211.72, ACP of 2.55\%) illustrates the inadequacy of directly applying general-purpose segmentation models to cell counting without task-specific tuning.

\begin{table}[h]
    \centering
    \scriptsize
        \caption{MAE across different cell types: AHPC and RPC. Lower values indicate better performance.}
    \label{tab:cell_type_performance}
    \begin{tabular}{l c c}
        \toprule
        \textbf{Model} & \textbf{AHPC} $\downarrow$ & \textbf{RPC} $\downarrow$ \\
        \midrule
       \rowcolor{Gray} \multicolumn{3}{l}{\textbf{Regression-based}} \\
        VGG-16 \cite{vgg16_liu} & \underline{39.23} & \underline{29.95} \\
        ResNet-50 \cite{He2015DeepRecognition} & 62.95 & 52.40 \\
        ResNet-18 \cite{He2015DeepRecognition} & 65.51 & 48.15 \\
        EfficientNet B7 \cite{efficientnet_tan} & 43.04 & 31.97 \\
        \midrule
        \rowcolor{Gray} \multicolumn{3}{l}{\textbf{Adopted Density Map Estimation from Crowd Counting}} \\
        MCNN \cite{zhang_single-image_2016} & 43.38 & 34.17 \\
        CSRNet \cite{li_csrnet_2018} & \underline{28.16} & 22.15 \\
        MAN \cite{lin2022boosting} & 28.84 & \underline{\textbf{18.88}} \\
        \midrule
        \rowcolor{Gray} \multicolumn{3}{l}{\textbf{Density Map Estimation for Cell Counting}} \\
        SAU-Net \cite{guo_sau-net_2019} & 43.10 & 33.83 \\
        C-FCRN+AUX \cite{he_deeply-supervised_2021} & 38.76 & 27.50 \\
        Count-ception \cite{paul_cohen_count-ception_2017} & 36.32 & \underline{23.20} \\
        FCRN-A \cite{xie_microscopy_2018} & \underline{32.06} & 31.21 \\
        \midrule
        \rowcolor{Gray} \multicolumn{3}{l}{\textbf{SAM-based Cell Counting}} \\
        IDCC-SAM (zero shot)~\cite{fanijo2025idcc} & 228.96 & 80.78 \\
        \textbf{SAM-Counter (Ours)} & \textbf{\underline{22.32}} & \underline{20.61} \\
        \bottomrule
    \end{tabular}
\end{table}

Table~\ref{tab:cell_type_performance} shows that SAM-Counter achieves the lowest error on AHPC images (22.32), demonstrating robust performance on data with high variability and sparsity. On RPC images, MAN slightly outperforms SAM-Counter with an MAE of 18.88 versus 20.61, suggesting that its architecture may be better suited for higher magnification images containing larger cells.

Among the baselines, CSRNet delivers competitive performance across both cell types, particularly on RPC (22.15), highlighting the general effectiveness of density map-based approaches. Count-ception and C-FCRN+AUX also exhibit notably lower error on RPC compared to AHPC, likely due to the reduced variability and tighter count distribution in the higher-magnification subset.

By contrast, regression-based models perform consistently worse on both conditions, with especially high errors on AHPC images, underscoring their limitations in handling spatial heterogeneity and large-scale count variation.

Overall, these results show that both model design choices and the nature of the dataset significantly impact performance in biological tasks.

\begin{table}[h]
    \centering
    \scriptsize
    \caption{Macro MAE ($\downarrow$) and ACP ($\uparrow$) for different cell density levels. Lower MAE and higher ACP indicate better performance. Underlined values are best within their respective categories; bold values are best overall.}
    \label{tab:cell_density_performance_combined}
    \resizebox{\linewidth}{!}{%
    \begin{tabular}{l c c c c c c}
        \toprule
        \multirow{2}{*}{\textbf{Model}} & 
        \multicolumn{2}{c}{\textbf{Low}} & 
        \multicolumn{2}{c}{\textbf{Medium}} & 
        \multicolumn{2}{c}{\textbf{High}} \\
        \cmidrule(lr){2-3} \cmidrule(lr){4-5} \cmidrule(lr){6-7}
        & MAE $\downarrow$ & ACP $\uparrow$ & MAE $\downarrow$ & ACP $\uparrow$ & MAE $\downarrow$ & ACP $\uparrow$ \\
        \midrule
        \rowcolor{Gray} \multicolumn{7}{l}{\textbf{Regression-based}} \\
        VGG-16 \cite{vgg16_liu} & \underline{13.70} & \underline{22.22} & 45.48 & \underline{32.35} & \underline{144.92} & \underline{30.23} \\
        ResNet-50 \cite{He2015DeepRecognition} & 25.92 & 11.11 & 78.93 & 8.82 & 212.97 & 16.28 \\
        ResNet-18 \cite{He2015DeepRecognition} & 22.77 & 13.13 & 56.11 & 23.53 & 256.80 & 9.30 \\
        EfficientNet B7 \cite{efficientnet_tan} & 14.45 & 20.20 & \underline{41.31} & 14.71 & 167.82 & 18.60 \\
        \midrule
        \rowcolor{Gray} \multicolumn{7}{l}{\textbf{Adopted Density Map Estimation from Crowd Counting}} \\
        MCNN \cite{zhang_single-image_2016} & 15.75 & 23.74 & 37.01 & 23.53 & 168.78 & 32.56 \\
        CSRNet \cite{li_csrnet_2018} & 7.70 & 38.38 & 25.54 & \underline{52.94} & \underline{119.99} & 27.91 \\
        MAN \cite{lin2022boosting} & \textbf{\underline{7.31}} & \underline{43.43} & \underline{23.82} & 47.06 & 124.53 & \underline{34.88} \\
        \midrule
        \rowcolor{Gray} \multicolumn{7}{l}{\textbf{Density Map Estimation for Cell Counting}} \\
        SAU-Net \cite{guo_sau-net_2019} & 17.62 & 22.22 & 41.56 & 29.41 & 154.75 & 20.93 \\
        C-FCRN+AUX \cite{he_deeply-supervised_2021} & \underline{10.29} & 30.81 & \underline{28.90} & \underline{44.12} & 169.25 & \underline{39.53} \\
        Count-ception \cite{paul_cohen_count-ception_2017} & 11.47 & \underline{36.87} & 31.84 & 35.29 & 144.49 & 25.58 \\
        FCRN-A \cite{xie_microscopy_2018} & 10.50 & 32.32 & 34.51 & 29.41 & \underline{128.78} & 34.88 \\
        \midrule
        \rowcolor{Gray} \multicolumn{7}{l}{\textbf{SAM for Cell Counting}} \\
        IDCC-SAM (zero shot)~\cite{fanijo2025idcc} & 45.10 & 2.02 & 152.47 & 5.88 & 1025.79 & 2.33 \\
        \textbf{SAM-Counter (Ours)} & \underline{9.17} & \textbf{\underline{47.47}} & \textbf{\underline{21.41}} & \textbf{\underline{58.82}} & \textbf{\underline{82.33}} & \textbf{\underline{46.51}} \\
        \bottomrule
    \end{tabular}}
\end{table}

\textbf{Performance across varying numbers of cells per image:}

We evaluate model performance over three cell-count ranges: \textit{low} (0–250 cells, 198 images), \textit{medium} (251–500 cells, 34 images), and \textit{high} (more than 500 cells, 43 images), based on ground-truth annotations.These ranges were defined in collaboration with domain biologists to reflect common variations in microscopy data. Table~\ref{tab:cell_density_performance_combined} reports macro MAE ($\downarrow$) and ACP ($\uparrow$) for each range.

SAM-Counter achieves the lowest MAE in the medium (21.41) and high (82.33) ranges and the highest ACP across all densities (47.47\%, 58.82\%, 46.51\%). Improvements are particularly clear in the high-density setting, where it reduces MAE by over 30\% compared to the next-best model (CSRNet, 119.99) and nearly doubles ACP (46.51\% vs. 27.91\%). For low-density images, SAM-Counter remains competitive (MAE 9.17) but is slightly outperformed by MAN (7.31) and CSRNet (7.70).

MAN achieves the best MAE in sparse images and a strong ACP (43.43\%), while CSRNet provides the best non-SAM high-density MAE (119.99) and the highest ACP at medium density (52.94\%). Among cell-specific DME models, C-FCRN+AUX achieves the lowest MAE for low and medium densities (10.29, 28.90) and the highest ACP within its group at high density (39.53\%).

Within model families, VGG-16 attains the best MAE for low (13.70) and high (144.92) densities, whereas EfficientNet-B7 is strongest at medium counts (41.31). MAN leads at low and medium densities among crowd-counting models, while CSRNet performs better at high density. FCRN-A is strongest for dense images (128.78), whereas C-FCRN+AUX and Count-ception perform better for sparse settings.

\subsection{Qualitative Analysis}
\begin{figure} [h]
  \centering
  \includegraphics[width=1\linewidth]{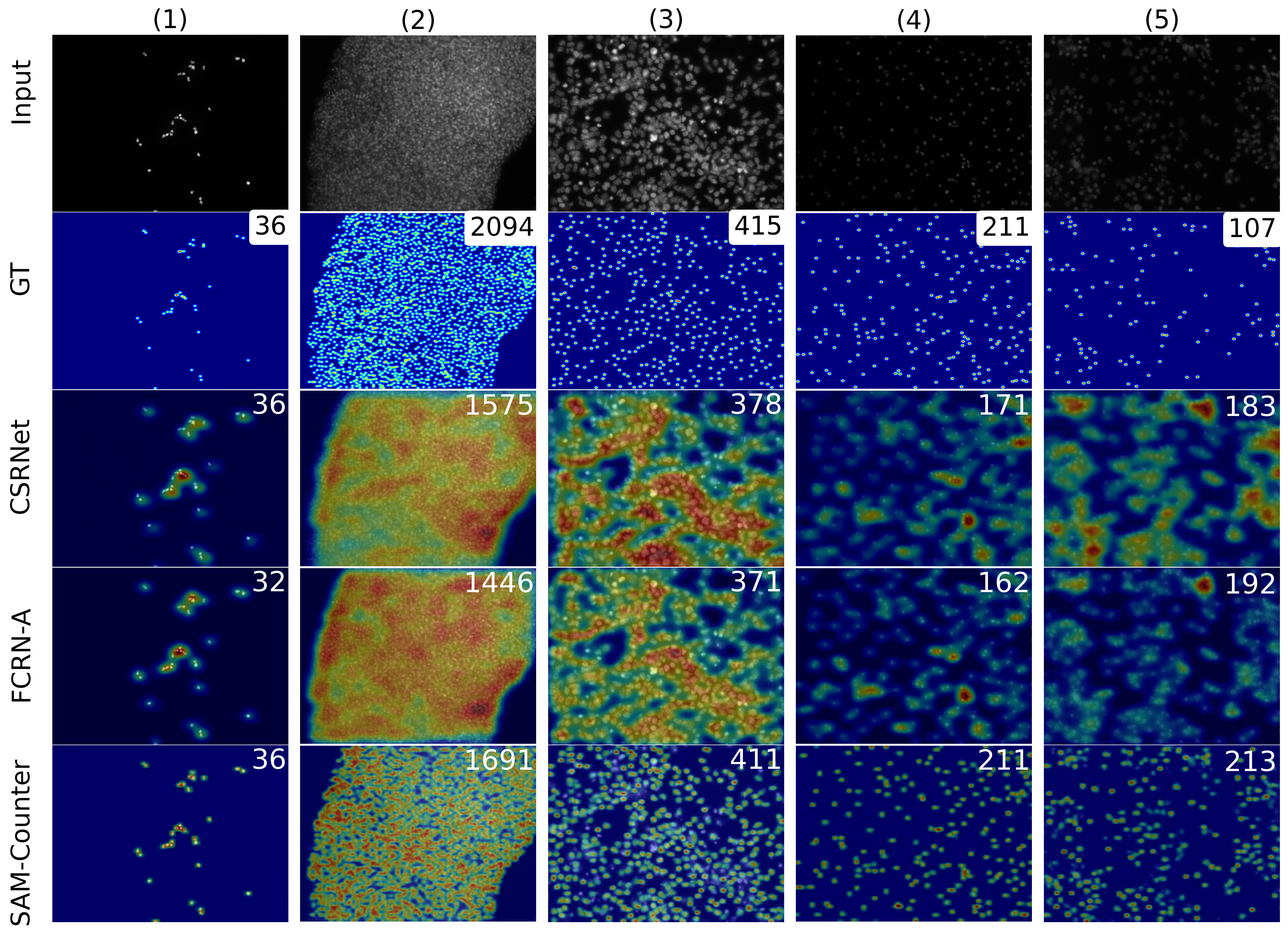} 
   \caption{Visualization of density map predictions of the three best-performing models on five representative test samples: one per column.} 
  \label{fig:preds}
\end{figure}

Fig.~\ref{fig:preds} shows qualitative results for the best-performing models selected based on the lowest Mean Absolute Error (MAE) from Table~\ref {tab:model_performance}: CSRNet from crowd counting methods, FCRN-A from cell counting methods, and SAM-Counter.  These models were evaluated on diverse test scenarios, ranging from sparse to highly congested scenes. 

The top row presents the input images, while the second row displays the ground truth density maps with the total number of cells indicated. Subsequent rows illustrate predicted density maps by CSRNet, FCRN-A, and SAM-Counter, respectively. Heatmap colors indicate density intensity, with red representing high density and blue indicating lower density regions.

Fig.~\ref{fig:preds}(1) illustrates a sparse scenario, where all three models show strong visual consistency with ground truth, accurately capturing individual cell locations. CSRNet and SAM-Counter precisely match the GT count, while FCRN-A slightly underestimates.
In Fig.~\ref{fig:preds}(2), a highly dense and challenging scenario with clustered cells is presented. All models produce smoother density maps, with CSRNet and FCRN-A significantly underestimating the count, whereas SAM-Counter demonstrates improved localization, though it slightly underestimates the total count.

Fig.~\ref{fig:preds}(3) shows moderately dense and variably sized cells. CSRNet and FCRN-A yield irregular density predictions, while SAM-Counter provides a uniform density distribution closely aligning with GT. Fig.~\ref{fig:preds}(4) represents a scenario with fewer cells and moderate density variations, where SAM-Counter closely matches the ground truth count and provides accurate cell localization, whereas CSRNet and FCRN-A slightly underestimate and produce more diffused density predictions.
Fig.~\ref{fig:preds}(5) depicts another sparse distribution scenario with varied cell sizes. SAM-Counter maintains consistent accuracy, closely mirroring GT distributions despite a slight overestimation. CSRNet and FCRN-A again exhibit challenges in precise localization and overrepresent overall density.

Overall, SAM-Counter tends to produce well-defined and visually consistent density maps across diverse scenarios, though all three models exhibit specific strengths and challenges depending on scenario complexity and cell distribution characteristics.

\subsection{Ablation Results}
\label{sec:ablation}

To evaluate the impact of design choices when adapting SAM, we investigated two configurations. One variable is whether to freeze or fine‑tune the SAM encoder. The other is the number of layers in the density estimation head. Table~\ref{tab:ablation_combined} reports the test MAE for each configuration.

\begin{table}[h]
  \centering
  \caption{Ablation results on SAM‐Counter: encoder freezing and density‐head depth (test MAE).}
  \label{tab:ablation_combined}
  \begin{tabular}{llcc}
    \toprule
    \textbf{Experiment}     & \textbf{Setting}      & \textbf{\# Params} & \textbf{Test MAE} \\
    \midrule
    \multirow{2}{*}{Encoder} 
                            & Frozen           & 41.2 K     & 44.24            \\
                            & Trainable        & 89.7 M  + 41.2 K   & 22.12            \\
    \addlinespace
    \multirow{3}{*}{Number of layers} 
                            & 1 layer           & 89.7 M + 257    & 22.99         \\
                            & 2 layers          & 89.7 M + 33K    & 24.58            \\
                            & 3 layers           & 89.7 M + 41.2 K   & 22.12            \\
    \bottomrule
  \end{tabular}
\end{table}

Table~\ref{tab:ablation_combined} shows the impact of encoder fine-tuning and head depth. Fine-tuning the SAM encoder reduces test MAE from 44.24 (frozen) to 22.12. This confirms that adapting the encoder to microscopy images improves performance.

The number of layers in the density estimation head also affects accuracy. A three-layer head performs best, with an MAE of 22.12. One and two layers result in slightly higher errors. The head remains lightweight in all cases. Even the largest version adds only about 41K parameters, compared to 89.7M in the encoder.

We evaluated whether SAM-Counter trained on DAPI-stained images could generalize to other staining types. The model was trained only on DAPI and tested on images stained with Cy3 and AF488, which correspond to different antibody markers. On these test images, SAM-Counter achieved MAE of 606.26. This performance drop supports our earlier observation: images stained with different markers show substantial visual differences (see Fig. \ref{fig:background}), which limit generalization across different fluorescence markers.

\subsection{Limitations and Broader Impacts}

While CellFMCount presents a diverse benchmark with multiple cell types, conditions, and magnifications, it does not cover all imaging modalities or biological settings. SAM-Counter is based on SAM~\cite{kirillov2023segany}, which requires 1024\texttimes1024 input images. Future work could explore parameter-efficient fine-tuning (e.g., LoRA) or model distillation to reduce resource requirements without sacrificing accuracy.

Our dataset and method aim to support research in neuroregeneration, cancer, and stem cell therapy by improving accuracy and consistency in cell counting. The data come from experiments on murine and rat cells. We demonstrate the potential of adapting foundation models for biomedical tasks and encourage responsible use and validation in downstream applications.

\section{Conclusion}
We introduced CellFMCount, a large-scale, diverse fluorescence microscopy dataset for robust cell counting under real-world variability. We showed how foundation models like SAM can be repurposed for density map estimation via our SAM-Counter, and benchmarked thirteen state-of-the-art methods on DAPI-stained images. Our results establish strong baselines and highlight both the challenges and opportunities in automated cell counting.

\bibliographystyle{unsrt} % or try abbrvnat or unsrtnat
\bibliography{references}
\end{document}